# A Survey on Sentence Embedding Models Performance for Patent Analysis

-- Preliminary Draft, Work in Progress --

**Hamid Bekamiri, Daniel S. Hain, Roman Jurowetzki**

Aalborg University Business School, Denmark

## Abstract

Patent data is an important source of knowledge for innovation research, while the technological similarity between pairs of patents is a key enabling indicator for patent analysis. Recently researchers have been using patent vector space models based on different NLP embeddings models to calculate the technological similarity between pairs of patents to help better understand innovations, patent landscaping, technology mapping, and patent quality evaluation. More often than not, Text Embedding is a vital precursor to patent analysis tasks. A pertinent question then arises: How should we measure and evaluate the accuracy of these embeddings? To the best of our knowledge, there is no comprehensive survey that builds a clear delineation of embedding models' performance for calculating patent similarity indicators. Therefore, in this study, we provide an overview of the accuracy of these algorithms based on patent classification performance and propose a standard library[1] and dataset[2] for assessing the accurcy of embeddings models based on PatentSBERTa approach. In a detailed discussion, we report the performance of the top 3 algorithms at section, class, and subclass levels. The results based on the first claim of patents show that PatentSBERTa, Bert-for-patents, and TF-IDF Weighted Word Embeddings have the best accuracy for computing sentence embeddings at the subclass level. According to the first results, the performance of the models in different classes varies, which shows researchers in patent analysis can utilize the results of this study to choose the best proper model based on the specific section of patent data they used.

---

[1] https://pypi.org/project/patent-classification/
[2] The code to preproduce the work presented here can be accessed at https://github.com/AI-Growth-Lab/SEMP_Patent. The dataset provided at https://huggingface.co/datasets/AI-Growth-Lab/patents_claims_1.5m_traim_test

# Introduction

Measuring patent to patent (p2p) similarity is the most effective way to distinguish the technological overlap, similarity, and technology landscaping. There are three main types of p2p similarity methods: citation based analysis, classification based analysis, and textual based methods. The citation based analysis uses backward and forward citations to calculate the similarity between a new patent to existing patents. The classification bassed one utilizes the co-classification matrix to evaluate the p2p similarity score. Textual based methods are based on natural language processing (NLP) that applies Machine Learning (ML) techniques to the textual components of patent data to derive p2p technological similarity indicators (Hain et al., 2022). All three approaches are subject to several limitations. The citation based approach focuses on prior works rather than technological content. In addition, many patents do not have any citation records, which leads to the inability to measure their similarity to other patents. Categorizing patents based on the classification approach is too broad for technology landscaping and patent retrieval, and the method fails to differentiate patents within a specific class or subclass (Hain et al., 2022; Younge and Kuhn, 2016). Although textual based analysis requires significant expertise in ML, NLP, information retrieval (IR), and data analysis, it achieves state-of-the-art results compared to other approaches. That is the reason why the majority of researchers have recently focused on applying this approach for calculating p2p similarity indicators.

In the context of ML, text embeddings are fundamentally a form of word or sentence representation by learned representations that prepare texts in an understandable format for a machine (Krestel, etal, 2021). Text embeddings map text data into vectors that can be a set of real numbers (a vector). Word or Sentence embedding is a term used for representing words or sentences in a text that encodes the meaning of the word or the sentence in n-dimensional space. It is expected for words or sentences that are closer in the vector space to be more similar. It is found from the related literature review that there exist three dominant word embeddings namely, Traditional word embedding, Static word embedding, Contextualized word embedding (Selva and Kanniga, 2021), and two-sentence embeddings approach, Non-Parameterized and Parameterized models (Wang and Kuo, 2020).

More often than not, Text Embedding is a vital precursor to patent analysis tasks. This while the language used in the patent is usually vague and filled with technical jargon. Therefore, extracting features can be challenging. Technical jargon, the use of legal language, and attention to intellectual property rights represent a challenge that can significantly affect the accuracy of patent analysis applications (Zhang et al., 2005; QI et al., 2020; Beal & Kafadar, 2008; Tseng et al., 2007). Thus, identifying appropriate embedding approaches to embed information accurately in patent texts seems to be necessary more than in other domains. A pertinent question then arises: How should we measure and evaluate the accuracy of these embeddings? Since there is a lack of a ground-truth benchmark dataset for an annotated p2p similarity, we cannot directly validate how accurate the embeddings are (Hain et al., 2022). Meanwhile, this limitation makes it difficult to obtain instructions for using appropriate embedding models in the field of patent analysis. Owing to the fact that using other approaches to calculate semantic similarities like citation based and classification based can be proper replacement metrics for evaluating the accuracy of embeddings, in this study, we use the PatentSBERTa approach

(Bekamiri et al., 2021) to assess the accuracy of different embedding models based on classification performance. So based on the performance of multi-label classification, we evaluate the accuracy of the embeddings.

This paper aims to provide an overview of the accuracy of the different types of sentence embedding models for patent analysis and propose a standard library and dataset for assessing the accurcy of embeddings models based on PatentSBERTa approach. The result of the study for the overview of the sentence embeddings accuracy will be reported on two levels, including the best models as a universal model for patent analysis in all sections and the best model for each section.

## Text Embeddings in Patent Analysis Tasks

There are various patent analysis tasks in which text information is an important source of knowledge for performing them e.g., calculating technological similarity between pairs of patents. The accuracy of text embedding plays an important role in these types of patent analysis tasks so that the precision of their results is directly dependent on the accuracy of embeddings models. The patent analysis tasks that text embeddings are considered the main pillar of them can be categorized into 6 groups (Shalaby and Zadrozny, 2019; Krestel et al, 2021). Table 1 illustrates the list of these patent analysis tasks which are detailed below.

**Patent Classification:** The patent classification is an automatic way of predicting the technical field of an new invention. At present, there are the two most popular patent classification systems include the International Patent Classification (IPC) and the Cooperative Patent Classification (CPC). In practical, IPC or CPC classification codes are assigend to a patent documents based on manual analysis. Thus, this procedure is a costly and time-consuming process. The patent classification task helps patent offices to do this process automaticaly in more effecient way. Since the introduction of the Transformer models for computing text embeddings, there has been a significant increase in using ML models based on text embeddings for patent classification (Hain et al., 2022; Bekamiri et al., 2021; Lee and Hasiang, 2020; Li et al., 2018; Shalaby et al., 2018). Most of these studies have focused on fine-tuning Transformer models for computing embeddings of abtracts, claims and titles of patent documents for predicting patent classification at the CPC subclass level. The experimental results of Roudsari et al. (2021) research show embeddings obtained from the Transformer model (GPT-2) outperform the previous models i.e., CBOW, Skip-gram, GloVe, and FastText.

**State of the Art Search (SOA):** SOA or Prior art search refers to comprehensive information retrieval from patent datasets, scientific literature, or any existing product. SOA is a recall task where a model focuses on retrieving all relevant documents at acceptable precision. Based on SOA all similar patents or publications to a given patent application are gathered in order to assess the novelty and inventiveness of an invitation. This procedure is often executed through keyword queries, which is time-consuming and error-prone. The purposes of SOA in different stages of the patent life cycle broadly can be categorized into 6 groups including Related work, Patentability, Infringement, Freedom to operate, Invalidity, and Technology survey (Shalaby and Zadrozny, 2019). For implementing all types of SOA, using text embeddings especially through transformer models have recently gained considerable attention (Bekamiri et al., 2021; Roudsari et al., 2021; Hofstätter et al., 2019). This while, by proposing new approaches for sentence embeddings through transformer models, the full-text search of a given patent application can be performed in which similar patents to a new patent application automatically be detected (Helmers et al., 2019) that significantly leads to improve accuracy and performance of the SOA.

**Patent Landscaping:** Patent landscaping aims to find related patents of a technical field or region, or of competitors and design a clear overview of the patent structure for assessing the trends in order to follow current trends in a specific technical field and avoid the risk of patent infringement (Choi et al., 2022). For patent landscaping, in the first step, all relevant patents must be extracted, which will be time-consuming and tedious if new ML-based approaches are not used. The main difference between patent landscaping and prior art search is in use cases and scale. Landscaping focuses on finding a large number of related patents for a subject. A prior art search, on the other hand, focuses on finding a small number of patents that are most similar to a particular patent. Text patent information including Claims, abstracts, and descriptions is the most precious information that can be used to patent landscaping. However, using this source of information via common traditional techniques such as a boolean keyword search has two significant limitations including polysemous complexity and constructing proper keyword queries for covering all aspects of the topic (Abood and Feltenberger, 2018). This while, through transformer models can exactly address these barriers. In contextual word embeddings, Transformer models by proper finetuning in target domain can capture the semantic relationships either when words have multiple meanings or when they would describe the same subject. In addition in symmetric semantic searches, instead of constructing a keyword query, a search can be performed through Transformer models by a seed set of patent texts such as abstracts or claims.

**Patent Quility and Market Valuation Analysis:** The patent quality analysis involves three related tasks: technological quality, legal quality, and commercial quality. The technological quality task is the fundamental step to patent quality analysis to evaluate the degree of nonobviousness that shows the impact of technical solution either for solving a difficult problem or generating unexpected technical effects. The legal quality task shows the writing quality of the application document to disclose the technology information and demand the claims. commercial quality refers to the market valuation which refers to the overall quality of patents as well (SONG Hefa1 , LI Zhenxing, 2021). Although the patent texts are crucial information for patent valuation, the majority of the studies just conducted simple statistical analyses based on patent citation networks. Some studies have focused on using Bag of words and LDA for utilizing patent text materials (Jin et al., 2014; Tang e al., 2012). The main limitation of these existing models is that they can not capture the relationships among words. In this regard, some recent scholars have proposed utilizing deep learning models to display these relationships (Chung and Sohn, 2020; Hsu et al., 2020; Lin et al., 2018). This while, transformer models as a significant paradigm shift for computing embeddings have been exactly proposed to address this limitation and able to prepare the conditions for using patent texts in patent valuation analysis.

**Technology Forecasting:** Technological forecasting (TF) is the study of new trends, the technologies, and timing of changes in technology (Roper et al., 2011). This task is a crucial reseach for goverments and enterprisers to help decision-makers for choosing the right direction in managerial issues such as resource allocation and priority setting in technology development. Given that innovations and emerging techinologis are published in the form of patent applications, patent data is an up-to-date source of knowledge for TF. There are varites type of approaches to implement TF studies, e.g., Cruve fitting, time-series model, etc. However majoitry of these approches utilize biblometric information and ignore rich text materials such as claims, abstracts, and descriptions in patent data. In this regard, some recent studies attepmted to use text analysis through deep learning for TF (Choi et al., 2022; Z. Qiu and Z. Wang, 2022; Lee and Kim, 2021; Zhou et al, 2020). While finding the most important patents are criual and precurcer task for some of these appraoches, transformer models can be effeiently concidered to detect relevant important patents in a specific technology field using patent text information.

**Patent Litigation Analysis:** Patent litigation is a law process where a patent holder leads to sue a company or individual over the unlawful use of a patent. In most cases usually, this procedure is a costly and time-consuming process. Predicting patent litigation based on machine learning can be a effective procative approach for reducing the relative cost of this process (Trappey et al., 2020). Although there are a few studies that attempt to use predictive analysis for patent litigation based on patent txet information (Trappey et al., 2020; Liu et al., 2018), the rapid growth of using transformer models for patent analysis task can be a effective factor for using more the text information in patent litigation.

| Patent Analysis Task | Purpose | Used in |
|---|---|---|
| Patent Classifcation | The patent classification is an automatic way of predicting the technical field of an new invention. In practical, IPC or CPC classification codes are assigend to a patent documents based on manual analysis. The patent classification task helps patent offices to do this process automaticaly in more effecient way. | Hain et al., 2022; Bekamiri et al., 2021; Lee and Hasiang, 2020; Li et al., 2018; Shalaby et al., 2018; Roudsari et al., 2021 |
| State of the art Search (SOA) | SOA or Prior art search refers to a comprehensive information retrieval from patent datasets, scientific literature or any existing product. SOA is a total recall task where a model focuses on retrieving all relevant documents at acceptable precision. | Shalaby and Zadrozny, 2019; Bekamiri et al., 2021; Roudsari et al., 2021; Hofstätter et al., 2019; Helmers et al., 2019 |
| Patent Landscaping | Patent landscaping aims to find related patents of a technical field or region, or of competitors and design a clear overview of the patent structure within them for assessing the trends in a technical field. | Hain et al., 2022; Choi et al., 2022; Abood and Feltenberger, 2018 |

| Patent Quility and Market Valuation Analysis | Patent quality analysis are involved three related tasks: technological quality, legal quality and commercial quality. Technological quality task is fundamental step to patent quality analysis to evaulate the degree of nonobviousness that shows the impact of technical solution either for solving a difficult problem or genereateing unexpected technical effect. Legal quality task shows the writing quality of the application document to disclose the tecnology information and demand the claims. commercial quality refers to the market valuation which refer to overall quility of patents as well. | Chung and Sohn, 2020; Hsu et al., 2020; Lin et al., 2018 |
|---|---|---|
| Technology Forecasting | Technological forecasting is the study of new trends, the technologies, and timing of changes in technology (Roper et al., 2011) | Choi et al., 2022; Z. Qiu and Z. Wang, 2022; Lee and Kim, 2021; Zhou et al, 2020 |
| Litigation Analysis | Patent litigation is a law process where a patent holder leads to sue a company or individual over the unlawful use of a patent. More often than not, this procedure is a costly and time-consuming process. | Trappey et al., 2020; Liu et al., 2018 |

Table 1: The patent analysis tasks that text embeddings is considered the main pillar of them (Shalaby and Zadrozny, 2019; Krestel et al, 2021)

## Text Embedding approaches in NLP

There exist three major word embedding techniques and two different sentence embedding approaches. The following subsections present surveys on these techniques.

### Word Embeddings Techniques

Based on the related literature review, the main word embedding techniques can be categorized into three groups, namely, Traditional, Static, and Contextualized word embedding.

Traditional word embedding is based on the frequency that considers the whole document and discovers the significance of rare words in the document, counts the occurrence of each word, and the co-occurrence of the words. Frequency-based embeddings fall into three catergories: Count vector, TF-IDF, and Co-occurrence. Count vector is a method of counting the times each word occurs and converting a given set of words into a frequency representation in the document. The main disadvantage of these techniques is that they cannot afford semantic information and similarity between two documents. The algorithm generates a vector with a length equal to the number of specified words. Vectors are calculated based on the frequency/counts of these words in the documents. TF-IDF is the abbreviation of Term Frequency-Inverse Document Frequency, defined as the number of times a term occurs in a document. In this method, TF gives more importance to the words appearing more frequently in a single document. From another point of view, IDF will try to diminish the weight of the words that occur multiple times across the other documents. Therefore the model identifies the main terms in the documents and ignores other uninformative terms at the same time. The co-occurrence matrix algorithm was based on the principle that similar words will occur concurrently and have similar contexts. The context window and Co-occurrence matrix are the crucial parts of this method in that the context window refers to the number of neighbors to a given word, and the occurrence matrix is built based on the number of times the two words in the context window occur together (Selva and Kanniga, 2021).

Static word embedding is based on predicting probabilities of the words and maps each word into a vector. These vectors are typically dense and have much lower dimensionality than the vocabulary size. The Static embeddings are trained by the lookup tables, which convert words into dense vectors. Word2Vec, published in 2013, is a method to construct text embeddings. This technique uses a neural network model to learn word representations from a sizable language resource consisting of a structured set of texts. It can be obtained using two methods (both involving Neural Networks): Skip Gram and Common Bag Of Words (CBOW). The difference between CBOW (Continuous Bag of Words) versus Skip-gram algorithms is that in the training in which the CBOW algorithm is used, the target word is given as output, and the neighboring words are given as input. In contrast, in Word2Vec training based on the Skip-Gram algorithm, the target word is inherently given as input, and the neighboring words are considered as output. Therefore, Word2Vec is trained by predicting the neighboring words from the target word as in the Skip-Gram algorithm, or the target word anticipates the adjourning words as it happens in the CBOW algorithm. The Glove is a technique where the model can be trained by the matrix factorization technique, which consists of the co-occurrence information of the words. The co-occurrence matrix gives information about the occurrence of the words in different pairs of words. The Glove model can be trained by the co-occurrence matrix, which can compute the ratio of probabilities between any two pairs of words. Bojanowski et al. (2016) proposed FastText, which is another word embedding method that represents each word as an n-gram of characters instead of using words directly like Word2Vec. It solves the main challenge of both Word2vec and GloVe as the limitation for computing vector representation of words that are not in the model dictionary (Bojanowski et al., 2017).

Although a word can have different semantic meanings in various sentences, in Traditional word embedding and Static word embeddings models, a word has a similar vector representation in disparate sentences. To address this challenge, Contextualized language models can capture these definitions learned from a large set of documents, so embedding a word will have different vector representations in different sentences. ELMo model (Peters et al., 2018) considers the entire input sentence for computing the word embeddings. Hence, an appropriate meaning, wherein a word could have multiple meanings or senses, can be recognized by the model. In addition, a word can have different ELMo vectors under different contexts. The Elmo word vectors are learned functions of the internal states of a deep bidirectional language model (biLM), which can be pre-trained on a large text corpus (Matthew et al., 2018). The third generation of Generative Pre-trained Transformer (GPT-3) is a neural network machine learning model, developed by OpenAI which trains by internet data to generate any type of text (Radford et al., 2018). The BERT model, which stands for Bidirectional Encoder Representations from Transformers, was introduced in 2018 by Google. BERT can be trained by jointly conditioning both the left and right sides of contexts in all layers from unlabeled texts. Therefore, it could be served in a specific domain without substantial architecture modifications (Devlin et al., 2018). Pre-trained Language Models (LMs) have recently become popular in NLP, as they empower transferring the learning in this domain. These models encode contextual information and high-level features of the language, modeling syntax and semantics, achieving state-of-the-art performance across a variety of tasks, such as named entity recognition (Peters et al., 2017), machine translation (Ramachandran et al., 2017) and text classification (Howard and Ruder, 2018). They are initially pre-trained on large corpora with high resource consumption and later fine-tuned on downstream tasks by adding one or more task-specific layers trained from scratch through in-domain data while other layers are frozen. Transformer models can have better results based on their architecture in specific tasks. In particular, GPT family models are mostly pre-trained for Machine translation, Text generation, Question and answering, and the BERT family for Document Classification and Regression tasks. Generally, fine-tuning allows the use of these models in a specific domain with little resource requirement and fewer data. However, using these models depends on the type of problem and the context in which they would be deployed.

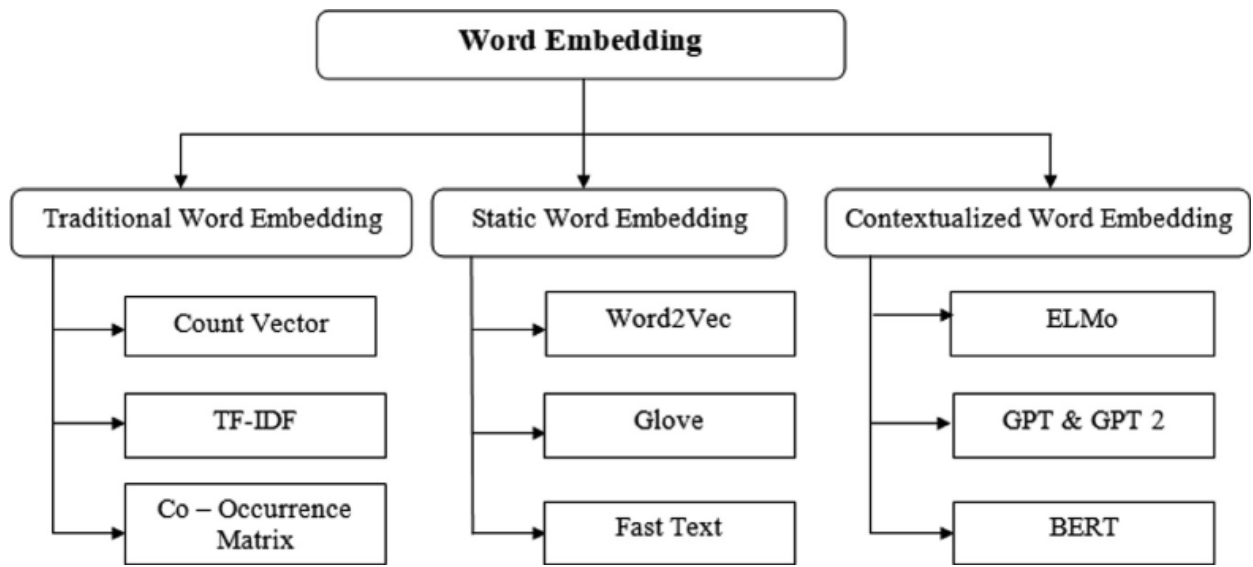

Fig. 1: Types of word embedding techniques (Selva and Kanniga, 2021)

**Sentence Embedding Approaches**

In the sentence embedding approach, instead of computing embedding at the word level, vectors of embeddings represent entire sentences and their semantic information. Generally, there are two main categories of sentence embedding approaches, from word embeddings to sentence embeddings, known as non-parameterized and parameterized models (Wang and Kuo, 2020).

Non-Parameterized methods are usually based on pre-trained word embedding methods, the most commonly used approach of which is to aggregate the embeddings of each word in a sentence using average-pooling, min-pooling, and max-pooling. Averaging (Avg) different word embeddings (Avg. GloVe and Avg. BERT) and weighted averaging methods (TF-IDF and smooth inverse frequency (SIF) (Arora, Liang, & Ma, 2019)) fall into this group.

Parameterized models utilize a more complex approach for computing sentence embeddings and usually perform better than non-parameterized models (SBERT-WK). In these models, acquiring the skip-thought model (Kiros et al., 2015) extends the unsupervised training of word embedding models (Mikolov et al., 2013) from the word level to the sentence one. The skip-thought model is a simple Neural Networks model for learning fixed-length representations of sentences in all NLP techniques without any labeled data. Recently parameterized models have attracted a lot of attention from different research areas and have had a lot of success. The most commonly used techniques in this regard are InferSent (Conneau et al., 2017), Universal Sentence Encoder, SBERT, and SBERT-WK. InferSent employs bidirectional Long Short-term Memory (LSTM) with supervised training.

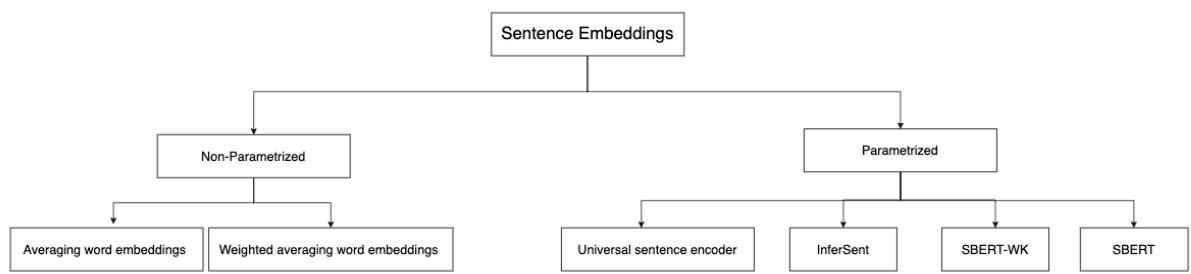

Fig. 2: Types of sentence embedding techniques

# Data

In this study, we used the EPO PATSTAT Worldwide Patent Statistical Database (2M patents for abstracts) and PatentsView dataset (1.5M patents for claims) provided by the US Patent and Trademark Office (USPTO). PATSTAT has been specifically developed for use by organizations and academic institutions. PatentsView is a patent data analysis platform designed to enhance the usefulness and transparency of US patent data and provides a regularly updated database connecting patents, inventors, organizations, and locations (USPTO, 2020). In this study, we utilized 2M patents randomly on EPO PATSTAT and 1.5 USPTO patents between 2013 and 2017, which at least have one claim on PatentsViews data source on the Google patent public datasets. The number of records available in this study was 3,492,294 patents

(1.5M of PatentView and 2M of PATSTAT), and we used 8 percent of the patents as the test dataset for evaluating the model. It is worth mentioning that we removed all duplicate records in patent id and claim text. Our exploration of data showed that we have about 163,909 different combinations of labels. For a better understanding of this point, Fig.3 shows the top 30 label combinations by value in the dataset.

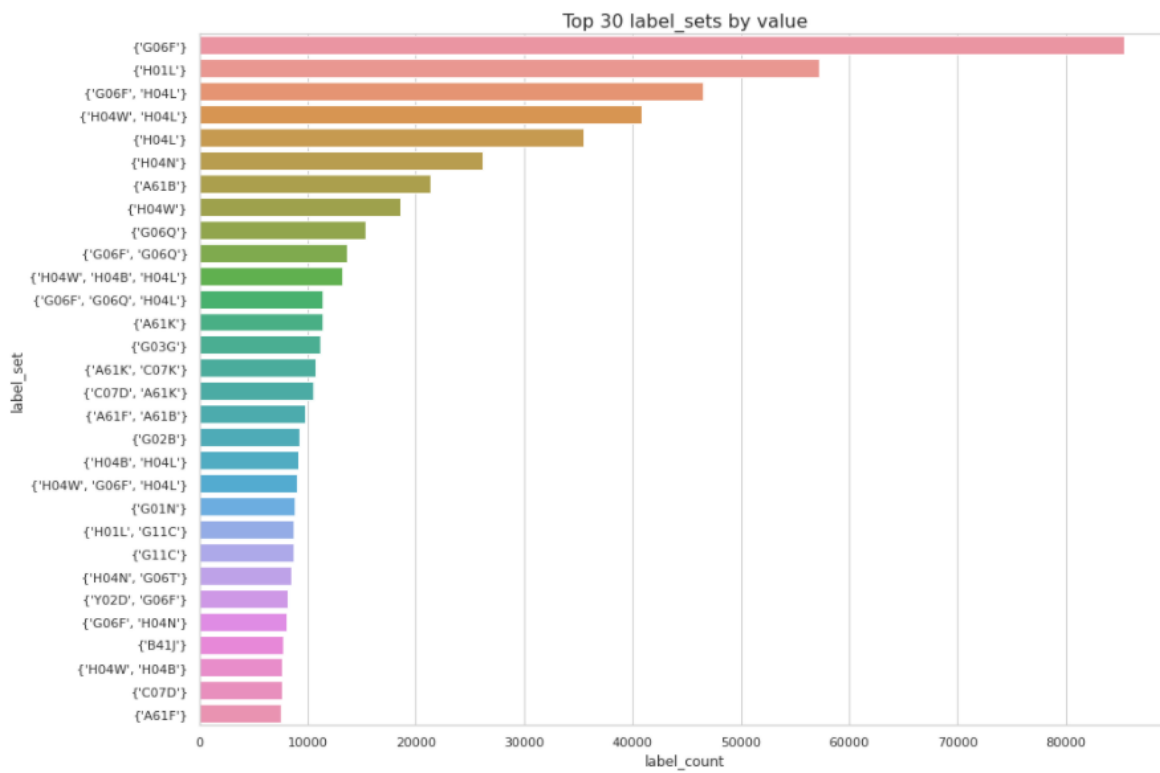

Fig. 3: Different combinations of the labels

## Computing different types of Sentence Embeddings

In this study, we evaluate the accuracy of one Non-Parameterized model as TF-IDF Weighted Word Embeddings (Hain et al., 2022) and four models of the Parameterized approach based on the SBERT model.

### Non-Parameterized: TF-IDF Weighted Word Embeddings

For this model, we trained a custom word2vec model on approximately 65 million English-language patent abstracts found in PATSTAT based on the specific language found in patent descriptions. Then we trained a TF-IDF model for a sample of the research dataset (3.5 M), which the TF-IDF model resulted in a matrix with 3.5M rows (patents) and 349790 columns (TF-IDF terms based on trained word2vec vocabulary). Subsequently, we computed the 300-dimensional trained word2vec embedding vector for each TF-IDF term. Finally, the sentence embeddings of the samples were computed by taking the dot-product of multiplying the word-embedding matrix with the dense TF-IDF weighted bag-of-words representations of the claims.

In this approach, we used a simple TF-IDF weighted approach by adding weights to each word based on its frequency within the document using the pre-trained word2vec (Lilleberg et al., 2015).

**Parameterized: SBERT**

In this study, based on Sentence Transformers, four different models were used to calculate p2p similarity by PatentSBERTa, Bert-for-patents, Bird-pegasus-large-bigpatent, and RoBERTa. Sentence Transformer is a framework based on PyTorch and Transformer models that can make this possible for researchers to implement a large sort of fine-tuned models for various tasks. The most important advantage of using this framework is that it makes it possible to propose a practical infrastructure to calculate p2p similarity with the performance of BERT.

PatentSBERTa is a pre-trained model based on SBERT that was fine-tuned through an augmented approach on 3432 pair claim selected sentences (Bekamiri et al., 2021). Bert-for-patents is a model based on BERT LARGE, trained by Google for 100M+ patents (not just US patents). Bird-pegasus-large-bigpatent was proposed and trained by a Google research group in 2021 that can handle sequences of length up to 8x of what was previously possible using similar hardware (Zaheer et al., 2021). RoBERTa is based on Google's BERT model released in 2018 and proposed in 2019. It modifies key hyperparameters, removing the next-sentence pretraining objective and training with much larger mini-batches and learning rates (Liu et al., 2019).

To compute the embeddings in this part, we only used the pre-trained models and did not consider any fine-tuning. Aforementioned, we used the Sentence Transformer framework and SBERT approach for computing sentence embeddings, and the sentence embeddings were calculated for 1.5 M claims and 2M abstracts of patents. In this study, we use only the first claim of the patents. Based on the architecture of SBERT models, we finally had a matrix of 1.5 M rows (claim of patents), 768 dimensions, 2M rows (abstract of patents), and 768 dimensions for every model.

# Evaluation

In this study, for evaluating embeddings predictive validity is used. Predictive validity is the extent to which a test measure can predict a criterion (Younge and Kuhn, 2016). In patent data, the clearest criterion measure is whether a pair of patents is also classified as similar by patent applicants and/or the USPTO. Two patents would be similar when they are similarly classified by the patent office, either explicitly through the USPTO's patent classification system or implicitly through assignment into the same administrative patent examining division (Younge and Kuhn, 2016). Given the previous statement, the performance of embedding-based patent classification models indicates the accuracy of embedding techniques, which indeed preserve the technological content and context of a patent (Hain et al., 2021).

Therefore, we used patent classification performance (PatntSBERTa paper approach) as a metric for evaluating the accuracy of the different types of sentence embedding techniques.

Based on the PatentSBERTa approach for multi-label patent classification and after finding similar semantic patents, we used a Sigmoid function for the prediction part, based on which we anticipated the labels of the instance.

$$Sigmoid\ (H) = \frac{1}{1 + e^{-x}}$$

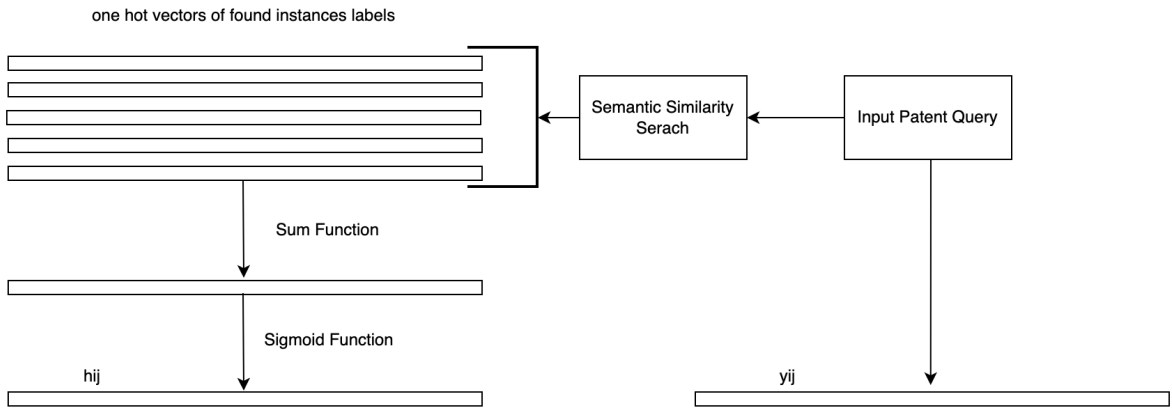

Fig. 4: Multi label prediction part based on semantic similarity results

As shown in Figure 4, labels of the found instances convert to one hot one, similar to the input query, then the number of times the labels appeared will be calculated. Finally, the sigmoid layer is used to classify the multi-label.

In this study, we used different metrics for reporting the performance of embedding models including Precision, Recall, Instance F1 score, Micro F1 score, Weighted F1 score, and Hamming Loss. The related formulas are reported as follows. In these formulas, assume yij the jth label of the ith instance, and yij = 1 or 0 means the jth label is relevant or irrelevant, respectively. H is the multi-label classifier, in which hji denotes the prediction of yij.

- Precision: Precision is the proportion of predicted correct labels to the total number of actual labels, averaged over all instances (Sorower, 2010).

$$Precision\ (H) = \frac{1}{m} \sum_{i=1}^{m} \frac{|Y_i \cap H_i|}{|H_i|}$$

- Recall: Recall is the proportion of predicted correct labels to the total number of predicted labels, averaged over all instances (Sorower, 2010).

$$Recall\,(H) = \frac{1}{m} \sum_{i=1}^{m} \frac{|Y_i \cap H_i|}{|Y_i|}$$

- Weighted F1-score: Calculate metrics for each label, and find their average weighted by support (the number of true instances for each label). This weight alters 'macro' to denote label imbalance; it can also result in an F-score that is not between precision and recall (Pedregosa et al.2011).

.
- F1 score (Instance-F1 score): This metric calculates the F1 score for each instance in the data and then calculates the average of the F1 scores (Xi-Zhu Wu and Zhi-Hua Zhou, 2017).

$$instance\text{-}F1(H) = \frac{1}{m} \sum_{i=1}^{m} \frac{2 \sum_{j=1}^{l} y_{ij} h_{ij}}{\sum_{j=1}^{l} y_{ij} + \sum_{j=1}^{l} h_{ij}}$$

- Micro-F1: Micro-F1 calculates metrics globally by counting the total true positives, false negatives, and false positives (Pedregosa et al.2011). Micro-F1 is a F-measure averaging on the prediction matrix.

$$micro\text{-}F1(H) = \frac{2 \sum_{j=1}^{l} \sum_{i=1}^{m} y_{ij} h_{ij}}{\sum_{j=1}^{l} \sum_{i=1}^{m} y_{ij} + \sum_{j=1}^{l} \sum_{i=1}^{m} h_{ij}}$$

- Hamming Loss: Hamming Loss reports the average of times the relevance of an example to a class label is incorrectly predicted (Schapire, 2000). Therefore, by predicting the incorrect label, the hamming loss represents the prediction error. In addition, where a relevant label is not anticipated, it identifies the missing error. It also normalizes over a total number of classes and a total number of examples.

$$hloss(H) = \frac{1}{ml} \sum_{i=1}^{m} \sum_{j=1}^{l} [\![ h_{ij} \neq y_{ij} ]\!]$$

where I is the indicator function. Ideally, we would expect the hamming loss as HL = 0, which implies no error. Practically the smaller the value of the hamming loss, the better the performance of the learning algorithm (Sorower, 2010).

## Results

In this section, we summarize our findings to compare the performance of different embedding models based on Non-parameterized and parameterized models. For the Non-parameterized group, we evaluated one model as TF-IDF Weighted Word Embeddings (Hain et al. 2022) and four models based on SBERT for the parameterized group. Table 1 shows the results of the models' performance assessed based on the claim and abstract of patents.

| Approach | Model | Claim | Abstract |
|---|---|---|---|
| Non-Parameterized | TF-IDF Weighted Word Embeddings (Hain et al, 2022) | 0.50 | 0.52 |
| | RoBERTa (Liu et al., 2019) | 0.40 | 0.44 |
| | Bert-for-patents (Srebrovic R and Yonamine J, 2020) | 0.65 | 0.66 |
| | Bird-pegasus-large-big patent (Zaheer et al., 2021) | 0.44 | 0.53 |
| Parameterized | PatentSBERTa (Bekamiri et al., 2021) | 0.68 | 0.69 |

Table 1: The accuracy of different embedding approaches based on patent multi-label classification performance at the subclass level

The performance of the embedding models shown in Table 1 illustrates the accuracy of well-known models like RoBERTa and Bird-pegasus-large-big-patent in sentence embeddings are not as good as traditional models like the TF-IDF Weighted Word Embedding model. There are two main reasons for the poor performance of RoBERTa and Bird-pegasus-large-big-patent: a lack of fine-tuning the model and the structure of the BERT models. The results show that while generally, RoBERT can improve for several supervised tasks, its performance is dependent on appropriate fine-tuning in the specific domain. In this study, due to the limitations in accessing the patent-labeled datasets, we did not execute fine-tuning the models.

On the other hand, Bird-pegasus-large-big-patent is a sparse-attention-based transformer that has extended the structure of Transformer models, therefore, it can handle larger contexts than 512 tokens as a common size of Transformer models. This feature made it handy for various NLP tasks such as summarization when we have a longer text, which is not the case in this study that the average claims and abstracts of the tokens are 162 words.

The result of Table 1 can be assumed as an evaluation of universal models (Choi et al. 2022) tested with about 100K instances in different contexts (sections). But we can check the performance result of sentence embedding models in specific sections shown in Table 2 and report the best performance of sentence embeddings in these sections. The result of this part of the study can help choose the best sentence embeddings model for researchers when they want to focus on a specific area and section of patents.

| Label | Explanation | RoBERTa | Bert-for-patents | PatentSBERTa | Best model |
|---|---|---|---|---|---|
| A | Human necessities | 0.68 | 0.81 | 0.80 | Bert-for-patents |
| B | Performing operations | 0.57 | 0.77 | 0.76 | Bert-for-patents |
| C | Chemistry | 0.79 | 0.84 | 0.86 | PatentSBERTa |
| D | Textiles | 0.17 | 0.63 | 0.64 | PatentSBERTa |
| E | Fixed constructions | 0.42 | 0.71 | 0.74 | PatentSBERTa |
| F | Mechanical engineering | 0.62 | 0.76 | 0.78 | PatentSBERTa |
| G | Physics | 0.77 | 0.85 | 0.85 | PatentSBERTa |
| H | Electricity | 0.79 | 0.88 | 0.86 | Bert-for-patents |
| Y | General tagging of new technological developments | 0.44 | 0.61 | 0.56 | Bert-for-patents |
| | Micro | 0.70 | 0.81 | 0.80 | |
| | Weighted | 0.69 | 0.81 | 0.80 | |
| | Instance | 0.72 | 0.83 | 0.82 | |

Table 2: The accuracy of selected embedding models based on patent multi-label classification performance at the section level

## Conclusion and Future work

We presented an overview of the sentence embeddings' accuracy for patent analysis. The idea is based on acquiring patent multi-label classification performance to evaluate the accuracy of the sentence embedding models. We reported the best universal embedding models for patent analysis as well as the most accurate model in every specific section of patent data.

We saw that the performance of some well-known BERT models like RoBERTa results in rather bad sentence embeddings, which is often worse than traditional models. The finding shows the importance of accessing a broader view of the accuracy of the embeddings models for patent analysis. In this regard, we compared several sentence embeddings models of the two groups known as parameterized and Non-parameterized.

The presented overview, which assesses the accuracy of the sentence embeddings is practical to select the most proper embedding model in patent analysis by researchers to embed information in patent texts accurately.

Based on this foundation, in the future work, we plan to explore the semi-supervised and unsupervised approaches for fine-tuning sentence embedding models and evaluate the effect of fine-tuning to improve the accuracy of these models.